\documentclass[11pt,a4paper]{article}
\usepackage[hyperref]{acl2021}
\usepackage{times}
\usepackage{latexsym}
\usepackage{multirow}
\usepackage{multicol}
\usepackage{makecell}
\usepackage{caption}
\usepackage{hyperref}
\usepackage{subcaption}
\usepackage{booktabs}
\usepackage{tikz}
\usepackage{xspace}
\usepackage{xargs}
\usepackage{amsmath}
\usepackage{amsfonts}
\usepackage{enumitem}

\DeclareMathOperator{\sign}{sign}
\DeclareMathOperator{\dist}{dist}
\DeclareMathOperator{\bert}{BERT}
\DeclareMathOperator{\softmax}{softmax}
\usepackage[T1]{fontenc}
\usepackage{microtype}
\newcommand*{\ditto}{\texttt{"}}

\aclfinalcopy % Uncomment this line for the final submission
\def\aclpaperid{1365} %  Enter the acl Paper ID here

\title{Efficient Passage Retrieval with Hashing for Open-domain\\Question Answering}

\author{
    Ikuya Yamada$^{\dagger,\ddagger}$\qquad
    Akari Asai$^*$\qquad
    Hannaneh Hajishirzi$^{*,\mathsection}$\\
    $^\dagger$Studio Ousia \quad
    $^\ddagger$RIKEN \quad
    $^*$University of Washington\\
    $^\mathsection$Allen Institute for AI \\\
    \texttt{ikuya@ousia.jp} \quad 
    \texttt{\{akari,hannaneh\}@cs.washington.edu} \\ 
}

\date{}

\begin{document}
\maketitle
\begin{abstract}
  Most state-of-the-art open-domain question answering systems use a neural retrieval model to encode passages into continuous vectors and extract them from a knowledge source. However, such retrieval models often require large memory to run because of the massive size of their passage index.
  In this paper, we introduce \textit{Binary Passage Retriever} (\textit{BPR}), a memory-efficient neural retrieval model that integrates a \textit{learning-to-hash} technique into the state-of-the-art Dense Passage Retriever (DPR) \cite{Karpukhin2020DenseAnswering} to represent the passage index using compact binary codes rather than continuous vectors.
  BPR is trained with a  multi-task objective over two tasks: efficient candidate generation based on binary codes and accurate reranking based on continuous vectors.
  Compared with DPR, BPR substantially reduces the memory cost from 65GB to 2GB without a loss of accuracy on two standard open-domain question answering benchmarks: Natural Questions and TriviaQA.
  Our code and trained models are available at \url{https://github.com/studio-ousia/bpr}.
\end{abstract}

\section{Introduction}
Open-domain question answering (QA) is the task of answering arbitrary factoid questions based on a knowledge source (e.g., Wikipedia).
Recent state-of-the-art QA models are typically based on a two-stage \textit{retriever--reader} approach  \cite{Chen2017ReadingQuestions} using a \textit{retriever} that obtains a small number of relevant passages from a large knowledge source and a \textit{reader} that processes these passages to produce an answer.
Most recent successful retrievers encode questions and passages into a common continuous embedding space using two independent encoders \cite{Lee2019LatentAnswering,Karpukhin2020DenseAnswering,Guu2020}.
Relevant passages are retrieved using a nearest neighbor search on the index containing the passage embeddings with a question embedding as a query.

These retrievers often outperform classical methods (e.g., BM25), but they incur a large memory cost due to the massive size of their passage index, which must be stored entirely in memory at runtime.
For example, the index of a common knowledge source (e.g., Wikipedia) requires dozens of gigabytes.\footnote{The passage index of the off-the-shelf DPR model \cite{Karpukhin2020DenseAnswering} requires 65GB for indexing the 21M English Wikipedia passages, which have 13GB storage size.}
A reduction in the index size is critical for real-world QA that requires large knowledge sources such as scientific databases (e.g., PubMed) and web-scale corpora (e.g., Common Crawl).

\begin{figure}[t]
  \centering
  \includegraphics[width=\linewidth]{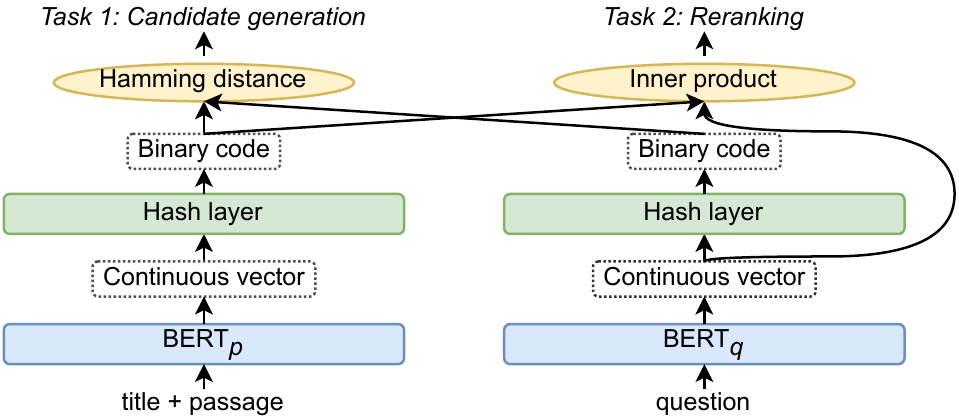}
  \caption{Architecture of BPR, a BERT-based model generating compact binary codes for questions and passages.
    The passages are retrieved in two stages: (1) efficient candidate generation based on the Hamming distance using the binary code of the question and (2) accurate reranking based on the inner product using the continuous embedding of the question.}
  \label{fig:architecture}
\end{figure}

In this paper, we introduce \textit{Binary Passage Retriever} (\textit{BPR}), which learns to hash continuous vectors into compact binary codes using a multi-task objective that simultaneously trains the encoders and hash functions in an end-to-end manner (see Figure \ref{fig:architecture}).
In particular, BPR integrates our \textit{learning-to-hash} technique into the state-of-the-art Dense Passage Retriever (DPR) \cite{Karpukhin2020DenseAnswering}
to drastically reduce the size of the passage index by storing it as binary codes.
BPR computes binary codes by applying the sign function to continuous vectors.
As the sign function is not compatible with back-propagation, we approximate it using the scaled tanh function during training.
To improve search-time efficiency while maintaining accuracy,  BPR is trained to obtain both binary codes and continuous embeddings for questions with multi-task learning over two tasks:
(1) candidate generation based on the Hamming distance using the binary code of the question and (2) reranking based on the inner product using the continuous embedding of the question.
The former task aims to detect a small number of candidate passages efficiently from the entire passages and the latter aims to rerank the candidate passages accurately.

We conduct experiments using the Natural Questions (NQ) \cite{doi:10.1162/tacl_a_00276} and TriviaQA (TQA) \cite{Joshi2017TriviaQA:Comprehension} datasets.
Compared with DPR, our BPR achieves similar QA accuracy and competitive retrieval performance with a substantially reduced memory cost from 65GB to 2GB.
Furthermore, using an improved reader, we achieve results that are competitive with those of the current state of the art in open-domain QA. 
Our code and trained models are available at \url{https://github.com/studio-ousia/bpr}.

\section{Related Work}
\paragraph{Retrieval for Open-domain QA}

Many recent open-domain QA models depend on the retriever to select relevant passages from a knowledge source.
Early works involved the adoption of sparse representations \cite{Chen2017ReadingQuestions} for the retriever, whereas recent works \cite{Lee2019LatentAnswering,Guu2020,Karpukhin2020DenseAnswering} have often adopted dense representations based on neural networks.
Our work is an extension of DPR \cite{Karpukhin2020DenseAnswering}, which has been used in recent state-of-the-art QA models \cite{Lewis2020Retrieval-augmentedTasks,Izacard2020LeveragingAnswering}.
Concurrent with our work, \citet{Izacard2020AAnswering} attempted to reduce the memory cost of DPR using post-hoc product quantization with dimension reduction and filtering of passages.
However, they observed a significant degradation in the QA accuracy compared with their full model.
We adopt the learning-to-hash method with our multi-task objective and substantially compress the index without losing accuracy.

\paragraph{Learning to Hash}
The objective of hashing is to reduce the memory and search-time cost of the nearest neighbor search by representing data points using compact binary codes.
Learning to hash \cite{Wang2016LearningSurvey,Wang2018AHash} is a method for learning hash functions in a data-dependent manner.
Recently, many \textit{deep-learning-to-hash} methods have been proposed \cite{Lai2015SimultaneousNetworks,Zhu2016DeepRetrieval,Li2016FeatureLabels,Cao2017HashNet:Continuation,Cao2018HashGAN:GAN} to jointly learn feature representations and hash functions in an end-to-end manner.
We follow \citet{Cao2017HashNet:Continuation} to implement our hash functions.
Similar to our work, \citet{Xu2020HashingSelection} used the learning-to-hash method to reduce the computational cost of the answer sentence selection task, the objective of which is to select an answer sentence from a limited number of candidates (up to 500 in their experiments).
Our work is different from the aforementioned work because we focus on efficient and scalable passage retrieval from a large knowledge source (21M Wikipedia passages in our experiments) using an effective multi-task approach.
In addition to hashing-based methods, improving approximate neighbor search has been actively studied \cite{Jegou2011ProductSearch,Malkov2020EfficientGraphs,Guo2020AcceleratingQuantization}. We use \citet{Jegou2011ProductSearch} and \citet{Malkov2020EfficientGraphs} as baselines in our experiments.

\section{Model}
Given a question and large-scale passage collection such as Wikipedia, a retriever finds relevant passages that are subsequently processed by a reader.
Our retriever is built on DPR \citep{Karpukhin2020DenseAnswering}, which is a retriever based on BERT \cite{devlin2018bert}.
In this section, we first introduce DPR and then explain our model.

\subsection{Dense Passage Retriever (DPR)}

DPR uses two independent BERT encoders to encode question $q$ and passage $p$ into $d$-dimensional continuous embeddings:
\begin{equation}
  \mathbf{e}_q = \bert_{q}(q),\,\,\, \mathbf{e}_p = \bert_{p}(p),
\end{equation}
where $\mathbf{e}_q \in \mathbb{R}^d$ and $\mathbf{e}_p \in \mathbb{R}^d$.
We use the uncased version of BERT-base; therefore, $d=768$.
The output representation of the \texttt{[CLS]} token is obtained from the encoder.
To create passage $p$, the passage title and text are concatenated (\texttt{[CLS]} \textit{title} \texttt{[SEP]} \textit{passage} \texttt{[SEP]}).
The relevance score of passage $p$, given question $q$, is computed using the inner product of the corresponding vectors, $\langle \mathbf{e}_q, \mathbf{e}_p \rangle$.

\paragraph{Training}
Let $\mathcal{D} = \{ \langle q_i, p^+_i, p^-_{i,1}, \cdots, p^-_{i,n} \rangle \}_{i=1}^m$ be $m$ training instances consisting of a question $q_i$, a passage that answers the question (positive passage), $p^+_i$, and $n$ passages that are irrelevant for the question (negative passages), $p^-_{i,j}$.
The model is trained by minimizing the negative log-likelihood of the positive passage:
\begin{equation}
  \resizebox{.89\hsize}{!}{
  $\mathcal{L}_{\text{dpr}} = -\log \frac{\exp(\langle \mathbf{e}_{q_i^{\phantom{0}}},\, \mathbf{e}_{p_i^+}\rangle)}{\exp(\langle \mathbf{e}_{q^{\phantom{0}}_i},\, \mathbf{e}_{p_i^+}\rangle) + \sum_{j=1}^n{\exp(\langle \mathbf{e}_{q^{\phantom{0}}_i},\, \mathbf{e}_{p^-_{i,j}}\rangle)}}$.
  }
  \label{eq:nll-loss}
\end{equation}

\paragraph{Inference}
DPR creates a passage index by applying the passage encoder to each passage in the knowledge source.
At runtime, it retrieves the top-$k$ passages employing maximum inner product search with the question embedding as a query.

\subsection{Model Architecture}
Figure \ref{fig:architecture} shows the architecture of BPR.
BPR builds a passage index by computing a binary code for each passage in the knowledge source.
To compute the binary codes for questions and passages, we add a \textit{hash layer} on top of the question and passage encoders of DPR.
Given embedding $\mathbf{e} \in \mathbb{R}^d$ computed by an encoder, the hash layer computes its binary code, $\mathbf{h} \in \{-1, 1\}^d$, as
\begin{equation}
  \mathbf{h} = \sign(\mathbf{e}),
\end{equation}
where $\sign(\cdot)$ is the sign function such that for $i = 1, ..., d$, $\sign(h_i) = 1$ if $h_i > 0$; otherwise, $\sign(h_i) = -1$.
However, the sign function is incompatible with back-propagation because its gradient is zero for all non-zero inputs and is ill-defined at zero.
Inspired by \citet{Cao2017HashNet:Continuation}, we address this by approximating the sign function using the scaled tanh function during the training:
\begin{equation}
  \mathbf{\tilde{h}} = \tanh(\beta \mathbf{e}),
\end{equation}
where $\beta$ is a scaling parameter.
When $\beta$ increases, the function gradually becomes non-smooth, and as $\beta\to\infty$, it converges to the sign function.
At each training step, we increase $\beta$ by setting $\beta = \sqrt{\gamma\cdot step + 1}$, where $step$ is the number of finished training steps.
We set $\gamma=0.1$ and explain the effects of changing it in Appendix \ref{sec:effects-gamma}.

\subsection{Two-stage Approach}
\label{subsec:approach}
To reduce the computational cost without losing accuracy, BPR splits the task into candidate generation and reranking stages.
At the candidate generation stage, we efficiently obtain the top-$l$ candidate passages using the Hamming distance between the binary code of question $\mathbf{h}_q$ and that of each passage, $\mathbf{h}_p$.
We then rerank the $l$ candidate passages using the inner product between the continuous embedding of question $\mathbf{e}_q$ and $\mathbf{h}_p$ and select the top-$k$ passages from the reranked candidates.
We perform candidate generation using binary code $\mathbf{h}_q$ for search-time efficiency, and reranking using expressive continuous embedding $\mathbf{e}_q$ for accuracy.
We set $l=1000$ and describe the effects of using different $l$ values in Appendix \ref{sec:effects-num-candidates}.

\begin{table*}[t]
  \centering
  \scalebox{0.82}{
    \begin{tabular}{l|cc|cc|cc|cc|c|c}
      \toprule
      \textbf{Model}                     & \multicolumn{2}{c|}{\textbf{Top 1}} & \multicolumn{2}{c|}{\textbf{Top 20}} & \multicolumn{2}{c|}{\textbf{Top 100}} & \multicolumn{2}{c|}{\textbf{QA Acc. (EM)}} & \multirow{2}{*}{\makecell{\textbf{Index}                                                     \\\textbf{size}}} & \multirow{2}{*}{\makecell{\textbf{Query}\\\textbf{time}}}\\
                                         & NQ & TQA & NQ & TQA & NQ & TQA & NQ & TQA & & \\
      \midrule
      DPR & \textbf{46.0} & \textbf{53.5} & 78.4 & \textbf{79.4} & 85.4 & \textbf{85.0} & 41.5 & \textbf{56.8} & 64.6GB  & 456.9ms \\
      DPR + HNSW & 45.7 & 53.2 & \textbf{78.8} & 78.8 & 85.2 & 84.2 & 41.2 & 56.6 & 151.0GB & 1.8ms   \\
      \midrule
      DPR + Simple LSH                   & 21.5                                & 28.4                                 & 63.9                                  & 65.2                                     & 77.2          & 76.9 & 35.8 & 48.1 & 2.0GB & 28.8ms\\
      DPR + PQ                           & 32.5                                & 42.8                                 & 72.2                                  & 73.2                                     & 81.2          & 80.4 & 38.4 & 52.0 & 2.0GB& 44.0ms \\
      \midrule
      BPR (linear scan; $l=1000$) & 41.1 & 49.7 & 77.9 & 77.9 & \textbf{85.7} & 84.5 & \textbf{41.6} & \textbf{56.8} & 2.0GB & 85.3ms \\
      BPR (hash table lookup; $l=1000$) & \ditto & \ditto & \ditto & \ditto & \ditto & \ditto & \ditto & \ditto & 2.2GB & 38.1ms \\
      \bottomrule
    \end{tabular}
  }
  \caption{Top $k$ recall and exact match (EM) QA accuracy on test sets with the index size and query time of BPR and baselines. All models use the same reader based on BERT-base to evaluate the QA accuracy.}
  \label{tb:retrieval-accuracy}
\end{table*}

\begin{table*}[t]
  \centering
  \scalebox{0.82}{
    \begin{tabular}{l|cc|cc|cc|c}
      \toprule
      \textbf{Model}                     & \multicolumn{2}{c|}{\textbf{Top 1}} & \multicolumn{2}{c|}{\textbf{Top 20}} & \multicolumn{2}{c|}{\textbf{Top 100}} &
      \multirow{2}{*}{\makecell{\textbf{Query}\\\textbf{time}}}\\
                                         & NQ                                  & TQA                                  & NQ                                    & TQA                                      & NQ            & TQA \\
      \midrule
      BPR ($l=1000$)       & 41.1                                & 49.7                                 & 77.9                                  & 77.9                                     & 85.7 & 84.5 & 38.1ms \\
      BPR w/o reranking    & 38.0                                & 46.1                                 & 76.5                                  & 75.9                                     & 84.9          & 83.4 & 37.9ms \\
      BPR w/o candidate generation               & 41.1                                & 49.7                                 & 77.9                                  & 77.9                                     & 85.7 & 84.5 & 457.8ms \\
      \bottomrule
    \end{tabular}
  }
    \caption{Results of our ablation study. Hash table lookup is used to implement candidate generation.}
  \label{tb:analysis}
\end{table*}

\begin{table*}[t]
  \centering
  \scalebox{0.82}{
    \begin{tabular}{l|c|c|cc}
      \toprule
      \textbf{Model}                                        & \textbf{Pretrained model}  & \textbf{\#params} & \textbf{NQ}   & \textbf{TQA}  \\
      \midrule
      RAG \cite{Lewis2020Retrieval-augmentedTasks} & BART-large & 406M   & 44.5          & 56.1          \\
      FiD (base) \cite{Izacard2020LeveragingAnswering} & T5-base & 220M & 48.2 &65.0 \\
      FiD (large) \cite{Izacard2020LeveragingAnswering} & T5-large & 770M     & \textbf{51.4} & \textbf{67.6} \\
      \midrule
      BPR ($l=1000$) & BERT-base &110M & 41.6 & 56.8 \\
      BPR ($l=1000$) & ELECTRA-large &335M & 49.0          & 65.6          \\
      \bottomrule
    \end{tabular}
  }
    \caption{Exact match QA accuracy of BPR and state of the art models. BPR achieves performance close to FiD (large) with almost half of the parameters.}
  \label{tb:e2e-qa}
\end{table*}

\subsection{Training}
\label{subsec:training}
To compute effective representations for both the candidate generation and reranking stages, we combine the loss functions of the two tasks:
\begin{equation}
  \mathcal{L} = \mathcal{L}_{\text{cand}} + \mathcal{L}_{\text{rerank}}.
\end{equation}

\paragraph{Task \#1 for Candidate Generation}
The objective of this task is to improve candidate generation using the ranking loss with the approximated hash code of question $\mathbf{\tilde{h}}_q$ and that of passage $\mathbf{\tilde{h}}_p$:
\begin{equation}
  \resizebox{.89\hsize}{!}{
  $\mathcal{L}_{\text{cand}} = \sum_{j=1}^{n}\max(0, -(\langle \mathbf{\tilde{h}}_{q_i^{\phantom{0}}}, \mathbf{\tilde{h}}_{p_i^+}\rangle + \langle \mathbf{\tilde{h}}_{q_i^{\phantom{0}}}, \mathbf{\tilde{h}}_{p_{i,j}^-}\rangle) + \alpha)$.
  }
  \label{eq:ranking-loss}
\end{equation}
We set $\alpha=2$ and investigate the effects of selecting different $\alpha$ values and using the cross-entropy loss instead of the ranking loss in Appendix \ref{sec:effects-binary-loss}.
Note that the retrieval performance based on the Hamming distance can be optimized using this loss function because the Hamming distance and inner product can be used interchangeably for binary codes.\footnote{Given two binary codes, $\mathbf{h}_i$ and $\mathbf{h}_j$, there exists a relationship between their Hamming distance, $\dist_H(\cdot,\cdot)$, and inner product, $\langle\cdot,\cdot\rangle$: $\dist_H(\mathbf{h}_i, \mathbf{h}_j) = \frac{1}{2}(const - \langle\mathbf{h}_i, \mathbf{h}_j\rangle)$.}

\paragraph{Task \#2 for Reranking}
We improve the reranking stage using the following loss function:
\begin{equation}
  \resizebox{.89\hsize}{!}{
  $\mathcal{L}_{\text{rerank}} = -\log \frac{\exp(\langle \mathbf{e}_{q_i^{\phantom{0}}},\, \mathbf{\tilde{h}}_{p_i^+}\rangle)}{\exp(\langle \mathbf{e}_{q^{\phantom{0}}_i},\, \mathbf{\tilde{h}}_{p_i^+}\rangle) + \sum_{j=1}^n{\exp(\langle \mathbf{e}_{q^{\phantom{0}}_i},\, \mathbf{\tilde{h}}_{p^-_{i,j}}\rangle)}}$.
  }
  \label{eq:reranking-loss}
\end{equation}
This function is equivalent to $\mathcal{L}_{\text{dpr}}$, with the exception that $\mathbf{\tilde{h}}_p$ is used instead of $\mathbf{e}_p$.

\subsection{Algorithms for Candidate Generation}
\label{subsec:inference}
To perform candidate generation, we test two standard algorithms: (1) linear scan based on efficient Hamming distance computation,\footnote{The Hamming distance can be computed more efficiently than the inner product using the POPCNT CPU instruction.} and (2) hash table lookup implemented by building a hash table that maps each binary code to the corresponding passages and querying it multiple times by increasing the Hamming radius until we obtain $l$ passages.

\section{Experiments}

\paragraph{Datasets}
We conduct experiments using the NQ and TQA datasets and English Wikipedia as the knowledge source.
We use the following preprocessed data available on the DPR website:\footnote{\url{https://github.com/facebookresearch/DPR}} Wikipedia corpus containing 21M passages and the training/validation datasets for the retriever containing multiple positive, \textit{random} negative, and \textit{hard} negative passages for each question.

\paragraph{Baselines}
We compare our BPR with DPR with linear scan and DPR with Hierarchical Navigable Small World (HSNW) graphs \cite{Malkov2020EfficientGraphs} -- which builds a multi-layer structure consisting of a hierarchical set of proximity graphs, following \citet{Karpukhin2020DenseAnswering} -- for our primary baselines.
We also apply two popular post-hoc quantization algorithms to the DPR passage index: simple locality sensitive hashing (LSH) \cite{Neyshabur2015OnSearch} and product quantization (PQ) \cite{Jegou2011ProductSearch}.
We configure these algorithms such that their passage representations have the same size as that of BPR: the number of bits per passage of the LSH is set as 768, and the number of centroids and the code size of the PQ are configured as 96 and 8 bits, respectively.

\paragraph{Experimental settings}
Our experimental setup follows \citet{Karpukhin2020DenseAnswering}.
We evaluate our model based on its top-$k$ recall (the percentage of positive passages in the top-$k$ passages), retrieval efficiency (the index size and query time), and exact match (EM) QA accuracy measured by combining our model with a reader.
We use the same BERT-based reader as that used by DPR.
Our model is trained using the same method as DPR.
We conduct experiments on servers with two Intel Xeon E5-2698 v4 CPUs and eight Nvidia V100 GPUs.
The passage index are built using Faiss \cite{Johnson2019Billion-scaleGPUs}.
Further details are provided in Appendix \ref{sec:detail-experimental-setup}.

\subsection{Results}
\paragraph{Main results}
Table \ref{tb:retrieval-accuracy} presents the top-$k$ recall (for $k \in \{1, 20, 100\}$), EM QA accuracy, index size, and query time achieved by BPR and baselines on the NQ and TQA test sets.
BPR achieves similar or even better performance than DPR in both retrieval with $k\geq20$ and EM accuracy with a substantially reduced index size from 65GB to 2GB.
We observe that BPR performs worse than DPR for $k=1$, but usually the recall in small $k$ is less important because the reader usually produces an answer based on $k \geq 20$ passages.
BPR significantly outperforms all quantization baselines.
The query time of BPR is substantially shorter than that of DPR.
Hash table lookup is faster than linear scan but requires slightly more storage.
DPR+HNSW is faster than BPR; however, it requires 151GB of storage.

\paragraph{Ablations}
Table \ref{tb:analysis} shows the results of our ablation study.
Disabling the reranking clearly degrades performance, demonstrating the effectiveness of our two-stage approach.
Disabling the candidate generation (treating all passages as candidates) results in the same performance as using only top-1000 candidates, but significantly increases the query time due to the expensive inner product computation over all passage embeddings.

\paragraph{Comparison with State of the Art}
\label{subsec:e2e-qa}
Table \ref{tb:e2e-qa} presents the EM QA accuracy of BPR combined with state-of-the-art reader models.
Here, we also report the results of our model using an improved reader based on ELECTRA-large \cite{Clark2020ELECTRA:Generators} instead of BERT-base.
Our improved model outperforms all models except the large model of Fusion-in-Decoder (FiD), which contains more than twice as many parameters as our model.

\section{Conclusion}
We introduce BPR, which is an extension of DPR, based on a learning-to-hash technique and a novel two-stage approach.
It reduces the computational cost of open-domain QA without a loss in accuracy.

\section*{Acknowledgement}
We are grateful for the feedback and suggestions from the anonymous reviewers and the members of the UW NLP group. This research was supported by Allen Distinguished investigator award, a gift from Facebook, and the Nakajima Foundation Fellowship.

\bibliographystyle{acl_natbib}
\bibliography{references}

\begin{thebibliography}{24}
\expandafter\ifx\csname natexlab\endcsname\relax\def\natexlab#1{#1}\fi

\bibitem[{Cao et~al.(2018)Cao, Liu, Long, and Wang}]{Cao2018HashGAN:GAN}
Yue Cao, Bin Liu, Mingsheng Long, and Jianmin Wang. 2018.
\newblock \href {https://doi.org/10.1109/CVPR.2018.00140} {{HashGAN: Deep
  Learning to Hash With Pair Conditional Wasserstein GAN}}.
\newblock In \emph{Proceedings of the IEEE Conference on Computer Vision and
  Pattern Recognition}, pages 1287--1296.

\bibitem[{Cao et~al.(2017)Cao, Long, Wang, and
  Yu}]{Cao2017HashNet:Continuation}
Z~Cao, M~Long, J~Wang, and P~S Yu. 2017.
\newblock \href {https://doi.org/10.1109/ICCV.2017.598} {{HashNet: Deep
  Learning to Hash by Continuation}}.
\newblock In \emph{2017 IEEE International Conference on Computer Vision},
  pages 5609--5618.

\bibitem[{Chen et~al.(2017)Chen, Fisch, Weston, and
  Bordes}]{Chen2017ReadingQuestions}
Danqi Chen, Adam Fisch, Jason Weston, and Antoine Bordes. 2017.
\newblock \href {https://doi.org/10.18653/v1/P17-1171} {{Reading Wikipedia to
  Answer Open-Domain Questions}}.
\newblock In \emph{Proceedings of the 55th Annual Meeting of the Association
  for Computational Linguistics (Volume 1: Long Papers)}, pages 1870--1879.

\bibitem[{Clark et~al.(2020)Clark, Luong, Le, and
  Manning}]{Clark2020ELECTRA:Generators}
Kevin Clark, Minh-Thang Luong, Quoc~V Le, and Christopher~D Manning. 2020.
\newblock {ELECTRA: Pre-training Text Encoders as Discriminators Rather Than
  Generators}.
\newblock In \emph{International Conference on Learning Representations}.

\bibitem[{Devlin et~al.(2019)Devlin, Chang, Lee, and
  Toutanova}]{devlin2018bert}
Jacob Devlin, Ming-Wei Chang, Kenton Lee, and Kristina Toutanova. 2019.
\newblock \href {https://doi.org/10.18653/v1/N19-1423} {{BERT: Pre-training of
  Deep Bidirectional Transformers for Language Understanding}}.
\newblock In \emph{Proceedings of the 2019 Conference of the North American
  Chapter of the Association for Computational Linguistics: Human Language
  Technologies, Volume 1 (Long and Short Papers)}, pages 4171--4186.

\bibitem[{Guo et~al.(2020)Guo, Sun, Lindgren, Geng, Simcha, Chern, and
  Kumar}]{Guo2020AcceleratingQuantization}
Ruiqi Guo, Philip Sun, Erik Lindgren, Quan Geng, David Simcha, Felix Chern, and
  Sanjiv Kumar. 2020.
\newblock \href {https://arxiv.org/abs/1908.10396} {{Accelerating Large-Scale
  Inference with Anisotropic Vector Quantization}}.
\newblock In \emph{International Conference on Machine Learning}.

\bibitem[{Guu et~al.(2020)Guu, Lee, Tung, Pasupat, and Chang}]{Guu2020}
Kelvin Guu, Kenton Lee, Zora Tung, Panupong Pasupat, and Mingwei Chang. 2020.
\newblock {Retrieval Augmented Language Model Pre-Training}.
\newblock In \emph{Proceedings of the 37th International Conference on Machine
  Learning}, volume 119, pages 3929--3938.

\bibitem[{Izacard and Grave(2020)}]{Izacard2020LeveragingAnswering}
Gautier Izacard and Edouard Grave. 2020.
\newblock {Leveraging Passage Retrieval with Generative Models for Open Domain
  Question Answering}.
\newblock \emph{arXiv preprint arXiv:2007.01282}.

\bibitem[{Izacard et~al.(2020)Izacard, Petroni, Hosseini, De~Cao, Riedel, and
  Grave}]{Izacard2020AAnswering}
Gautier Izacard, Fabio Petroni, Lucas Hosseini, Nicola De~Cao, Sebastian
  Riedel, and Edouard Grave. 2020.
\newblock {A Memory Efficient Baseline for Open Domain Question Answering}.
\newblock \emph{arXiv preprint arXiv:2012.15156}.

\bibitem[{J{\'{e}}gou et~al.(2011)J{\'{e}}gou, Douze, and
  Schmid}]{Jegou2011ProductSearch}
H~J{\'{e}}gou, M~Douze, and C~Schmid. 2011.
\newblock \href {https://doi.org/10.1109/TPAMI.2010.57} {{Product Quantization
  for Nearest Neighbor Search}}.
\newblock \emph{IEEE Transactions on Pattern Analysis and Machine
  Intelligence}, 33(1):117--128.

\bibitem[{Johnson et~al.(2019)Johnson, Douze, and
  J{\'{e}}gou}]{Johnson2019Billion-scaleGPUs}
J~Johnson, M~Douze, and H~J{\'{e}}gou. 2019.
\newblock \href {https://doi.org/10.1109/TBDATA.2019.2921572} {{Billion-Scale
  Similarity Search with GPUs}}.
\newblock \emph{IEEE Transactions on Big Data}.

\bibitem[{Joshi et~al.(2017)Joshi, Choi, Weld, and
  Zettlemoyer}]{Joshi2017TriviaQA:Comprehension}
Mandar Joshi, Eunsol Choi, Daniel Weld, and Luke Zettlemoyer. 2017.
\newblock \href {https://doi.org/10.18653/v1/P17-1147} {{TriviaQA: A Large
  Scale Distantly Supervised Challenge Dataset for Reading Comprehension}}.
\newblock In \emph{Proceedings of the 55th Annual Meeting of the Association
  for Computational Linguistics (Volume 1: Long Papers)}, pages 1601--1611.

\bibitem[{Karpukhin et~al.(2020)Karpukhin, Oguz, Min, Lewis, Wu, Edunov, Chen,
  and Yih}]{Karpukhin2020DenseAnswering}
Vladimir Karpukhin, Barlas Oguz, Sewon Min, Patrick Lewis, Ledell Wu, Sergey
  Edunov, Danqi Chen, and Wen-tau Yih. 2020.
\newblock \href {https://doi.org/10.18653/v1/2020.emnlp-main.550} {{Dense
  Passage Retrieval for Open-Domain Question Answering}}.
\newblock In \emph{Proceedings of the 2020 Conference on Empirical Methods in
  Natural Language Processing}, pages 6769--6781.

\bibitem[{Kwiatkowski et~al.(2019)Kwiatkowski, Palomaki, Redfield, Collins,
  Parikh, Alberti, Epstein, Polosukhin, Devlin, Lee, Toutanova, Jones, Kelcey,
  Chang, Dai, Uszkoreit, Le, and Petrov}]{doi:10.1162/tacl_a_00276}
Tom Kwiatkowski, Jennimaria Palomaki, Olivia Redfield, Michael Collins, Ankur
  Parikh, Chris Alberti, Danielle Epstein, Illia Polosukhin, Jacob Devlin,
  Kenton Lee, Kristina Toutanova, Llion Jones, Matthew Kelcey, Ming-Wei Chang,
  Andrew~M Dai, Jakob Uszkoreit, Quoc Le, and Slav Petrov. 2019.
\newblock \href {https://doi.org/10.1162/tacl{\_}a{\_}00276} {{Natural
  Questions: A Benchmark for Question Answering Research}}.
\newblock \emph{Transactions of the Association for Computational Linguistics},
  7:453--466.

\bibitem[{Lai et~al.(2015)Lai, Pan, Liu, and Yan}]{Lai2015SimultaneousNetworks}
Hanjiang Lai, Yan Pan, Ye~Liu, and Shuicheng Yan. 2015.
\newblock \href {https://doi.org/10.1109/CVPR.2015.7298947} {{Simultaneous
  Feature Learning and Hash Coding With Deep Neural Networks}}.
\newblock In \emph{Proceedings of the IEEE Conference on Computer Vision and
  Pattern Recognition}.

\bibitem[{Lee et~al.(2019)Lee, Chang, and Toutanova}]{Lee2019LatentAnswering}
Kenton Lee, Ming-Wei Chang, and Kristina Toutanova. 2019.
\newblock \href {https://doi.org/10.18653/v1/P19-1612} {{Latent Retrieval for
  Weakly Supervised Open Domain Question Answering}}.
\newblock In \emph{Proceedings of the 57th Annual Meeting of the Association
  for Computational Linguistics}, pages 6086--6096.

\bibitem[{Lewis et~al.(2020)Lewis, Perez, Piktus, Petroni, Karpukhin, Goyal,
  K{\"{u}}ttler, Lewis, Yih, Rockt{\"{a}}schel, and
  {others}}]{Lewis2020Retrieval-augmentedTasks}
Patrick Lewis, Ethan Perez, Aleksandara Piktus, Fabio Petroni, Vladimir
  Karpukhin, Naman Goyal, Heinrich K{\"{u}}ttler, Mike Lewis, Wen-tau Yih, Tim
  Rockt{\"{a}}schel, and {others}. 2020.
\newblock {Retrieval-Augmented Generation for Knowledge-Intensive NLP Tasks}.
\newblock In \emph{Advances in Neural Information Processing Systems 33}.

\bibitem[{Li et~al.(2016)Li, Wang, and Kang}]{Li2016FeatureLabels}
Wu-Jun Li, Sheng Wang, and Wang-Cheng Kang. 2016.
\newblock {Feature Learning Based Deep Supervised Hashing with Pairwise
  Labels}.
\newblock In \emph{Proceedings of the Twenty-Fifth International Joint
  Conference on Artificial Intelligence}, pages 1711--1717.

\bibitem[{Malkov and Yashunin(2020)}]{Malkov2020EfficientGraphs}
Y~A Malkov and D~A Yashunin. 2020.
\newblock \href {https://doi.org/10.1109/TPAMI.2018.2889473} {{Efficient and
  Robust Approximate Nearest Neighbor Search Using Hierarchical Navigable Small
  World Graphs}}.
\newblock \emph{IEEE Transactions on Pattern Analysis and Machine
  Intelligence}, 42(4):824--836.

\bibitem[{Neyshabur and Srebro(2015)}]{Neyshabur2015OnSearch}
Behnam Neyshabur and Nathan Srebro. 2015.
\newblock {On Symmetric and Asymmetric LSHs for Inner Product Search}.
\newblock In \emph{Proceedings of the 32nd International Conference on Machine
  Learning}, volume~37, pages 1926--1934.

\bibitem[{Wang et~al.(2016)Wang, Liu, Kumar, and
  Chang}]{Wang2016LearningSurvey}
J~Wang, W~Liu, S~Kumar, and S~Chang. 2016.
\newblock \href {https://doi.org/10.1109/JPROC.2015.2487976} {{Learning to Hash
  for Indexing Big Data—A Survey}}.
\newblock \emph{Proceedings of the IEEE}, 104(1):34--57.

\bibitem[{Wang et~al.(2018)Wang, Zhang, Song, Sebe, and Shen}]{Wang2018AHash}
J~Wang, T~Zhang, J~Song, N~Sebe, and H~T Shen. 2018.
\newblock \href {https://doi.org/10.1109/TPAMI.2017.2699960} {{A Survey on
  Learning to Hash}}.
\newblock \emph{IEEE Transactions on Pattern Analysis and Machine
  Intelligence}, 40(4):769--790.

\bibitem[{Xu and Li(2020)}]{Xu2020HashingSelection}
Dong Xu and Wu-Jun Li. 2020.
\newblock \href {https://doi.org/10.1609/aaai.v34i05.6473} {{Hashing Based
  Answer Selection}}.
\newblock \emph{Proceedings of the AAAI Conference on Artificial Intelligence},
  34(05):9330--9337.

\bibitem[{Zhu et~al.(2016)Zhu, Long, Wang, and Cao}]{Zhu2016DeepRetrieval}
Han Zhu, Mingsheng Long, Jianmin Wang, and Yue Cao. 2016.
\newblock {Deep Hashing Network for Efficient Similarity Retrieval}.
\newblock \emph{Proceedings of the AAAI Conference on Artificial Intelligence},
  30(1):2415–2421.

\end{thebibliography}

\appendix
\clearpage

\section*{Appendix for ``Efficient Passage Retrieval with Hashing for Open-domain Question Answering''}

\section{Details of Experimental Setup}
\label{sec:detail-experimental-setup}

\subsection{Knowledge Source}
As the knowledge source,  we use the preprocessed Wikipedia corpus consisting of 21,015,324 Wikipedia passages available on the website of \citet{Karpukhin2020DenseAnswering}.
The corpus is based on the December 20, 2018 version of the English Wikipedia and created by filtering out semi-structured data (i.e., tables, infoboxes, lists, and disambiguation pages) and splitting the remaining Wikipedia articles into multiple, disjointed text blocks of 100 words each.

\subsection{Question Answering Datasets}
\label{sec:dataset-datails}
We conduct experiments using the NQ and TQA datasets with the training, development, and test sets as in \citet{Lee2019LatentAnswering,Karpukhin2020DenseAnswering}.
A brief description of these datasets is provided as follows:

\begin{itemize}[leftmargin=10pt,topsep=1pt,itemsep=0pt]
\item \textbf{NQ} is a QA dataset for which questions are obtained from Google queries and answers comprise the spans of English Wikipedia articles.
\item \textbf{TQA} consists of trivia questions and their answers retrieved from the Web.
\end{itemize}

We use the preprocessed datasets available on the website of \citet{Karpukhin2020DenseAnswering}.\footnote{\url{https://github.com/facebookresearch/DPR}}
The numbers of questions contained in these datasets are listed in Table \ref{tb:dataset-details}.
For each question, the dataset contains three types of passages: (1) \textit{positive passages} selected based on gold-standard human annotations or distant supervision, (2) \textit{random negative passages} selected randomly from all the passages, and (3) \textit{hard negative passages} selected based on the BM25 scores between the question and all the passages.

\subsection{Details of BPR}
\label{details-training}
Our training configuration follows that of \citet{Karpukhin2020DenseAnswering}.
In particular, for each question, we use one positive and one hard negative passage and create a mini-batch comprising 128 questions.
We use the method of \textit{inbatch-negatives}, wherein each positive passage in a mini-batch is treated as the negative passage of each question in the mini-batch if it does not correspond to the question.
Our model contains 220 million parameters, and is trained for up to 40 epochs using Adam.
Regarding the hyperparameter search, we select the learning rate from the search range \{1e-5, 2e-5, 3e-5, 5e-5\} based on the top-100 recall on the validation set of the NQ dataset.
Therefore, the number of hyperparameter search trials is 4.
The detailed hyperparameters are listed in Table \ref{tb:biencoder-hyper-params}.

\begin{table}[t]
  \centering
  \small{
    \begin{tabular}{l|ccc}
      \toprule
      \textbf{Dataset}  & \textbf{Train} & \textbf{Validation} & \textbf{Test} \\
      \midrule
      NQ & 58,880         & 8,757                & 3,610         \\
      TQA          & 60,413         & 8,837                & 11,313        \\
      \bottomrule
    \end{tabular}
  }
    \caption{Number of questions in the preprocessed dataset used in our experiments.}
  \label{tb:dataset-details}
\end{table}

\begin{table}[t]
  \centering
  \small{
    \begin{tabular}{l|c}
      \toprule
      \textbf{Name}                    & \textbf{Value}  \\
      \midrule
      Batch size              & 128    \\
      Maximum question length & 256    \\
      Maximum passage length  & 256    \\
      Maximum training epochs & 40 \\
      Peak learning rate      & 2e-5   \\
      Learning rate decay     & linear \\
      Warmup ratio            & 0.06   \\
      Dropout                 & 0.1    \\
      Weight decay            & 0.0    \\
      Adam $\beta_1$          & 0.9    \\
      Adam $\beta_2$          & 0.999  \\
      Adam $\epsilon$         & 1e-6   \\
      \bottomrule
    \end{tabular}
  }
  \caption{Hyperparameters used to train BPR.}
  \label{tb:biencoder-hyper-params}
\end{table}

\subsection{Details of Reader}
\label{sec:reader}

For each passage in the top-$k$ passages retrieved by the retriever, the reader assigns a relevance score to the passage and selects the best answer span in the passage.
The final answer is the selected span from the passage with the highest relevance score.

Let $\mathbf{P}_{i} \in \mathbb{R}^{q \times d}$ ($1 \leq i \leq k$) be a BERT output representation for the $i$-th passage, where $q$ is the maximum token length of the passage, and $d$ is the dimension size of the output representation.
The probabilities of a passage being selected and  a token being the start or end positions of an answer is computed as

\begin{eqnarray}
  P_{score}(i) &=& \softmax\big( \mathbf{\hat{P}}^{\top} \mathbf{w}_{score}  \big)_i,\\
  P_{start,i}(s)  &=& \softmax \big(\mathbf{P}_{i} \mathbf{w}_{start}\big)_s, \\
  P_{end,i}(t) &=& \softmax \big(\mathbf{P}_{i} \mathbf{w}_{end}\big)_t,
\end{eqnarray}
where $\mathbf{\hat{P}} = [\mathbf{P}_1^{\mathrm{[CLS]}}, \ldots, \mathbf{P}_k^{\mathrm{[CLS]}}] \in \mathbb{R}^{d \times k}$, $\mathbf{w}_{score} \in \mathbb{R}^{d}$, $\mathbf{w}_{start} \in \mathbb{R}^{d}$, and $\mathbf{w}_{end} \in \mathbb{R}^{d}$.
The passage selection score of the $i$-th passage is given as $P_{score}(i)$, and the score of the $s$-th to $t$-th tokens from the $i$-th passage is given as $P_{{start},i}(s) \times P_{{end},i}(t)$.

During the training, we sample one positive and multiple negative passages from the passages returned by the retriever.
The model is trained to maximize the log-likelihood of the correct answer span in the positive passage,  combined with the log-likelihood of the positive passage being selected.
We use the BERT-base or ELECTRA-large as our pretrained model.
Regarding the hyperparameter search, we select the learning rate from \{1e-5, 2e-5, 3e-5, 5e-5\} based on its EM accuracy on the validation set of the NQ dataset.
Therefore, the number of hyperparameter search trials is 4.
Detailed hyperparameters are listed in Table \ref{tb:reader-hyper-params}.

\begin{table}[t]
  \centering
  \small{
  \setlength{\tabcolsep}{2pt}
    \begin{tabular}{l|cc}
      \toprule
      \textbf{Name}                    & \textbf{BERT-base} & \textbf{ELECTRA-large} \\
      \midrule
      Batch size            & 32     & 32  \\
      Maximum token length  & 350    & 350  \\
      Maximum training epochs & 20     & 20 \\
      Negative passage size & 23     & 17  \\
      Peak learning rate    & 2e-5   & 1e-5  \\
      Learning rate decay   & linear & linear\\
      Warmup ratio          & 0.06   & 0.06  \\
      Dropout               & 0.1    & 0.1   \\
      Weight decay          & 0.0    & 0.0   \\
      Adam $\beta_1$        & 0.9    & 0.9   \\
      Adam $\beta_2$        & 0.999  & 0.999 \\
      Adam $\epsilon$       & 1e-6   & 1e-6 \\
      \bottomrule
    \end{tabular}
  }
  \caption{Hyperparameters used to train the reader based on BERT-base and that based on ELECTRA-large.}
  \label{tb:reader-hyper-params}
\end{table}

\begin{table}[t]
  \centering
  \small{
    \begin{tabular}{l|ccc}
      \toprule
      \textbf{Configuration} & \textbf{Top 1} & \textbf{Top 20} & \textbf{Top 100} \\
      \midrule
      $\gamma=0.025$         & 39.4 & \textbf{76.7} & 83.8 \\
      $\gamma=0.05$          & 39.5 & 76.5 & 84.0 \\
      $\gamma=0.1$           & \textbf{39.8} & \textbf{76.7} & \textbf{84.1} \\
      $\gamma=0.2$           & 39.6 & 76.3 & 83.9 \\
      \bottomrule
    \end{tabular}
    \caption{Top-1, top-20, and top-100 recall of our model with $\gamma \in \{0.025, 0.05, 0.1, 0.2\}$ on the validation set of the NQ dataset.}
    \label{tb:experiments-gamma}
  }
\end{table}

\begin{table}[t]
  \centering
  \setlength{\tabcolsep}{5pt}
  \small{
    \begin{tabular}{l|cc|cc|cc}
      \toprule
      \textbf{\#candidates}                     & \multicolumn{2}{c|}{\textbf{Top 1}} & \multicolumn{2}{c|}{\textbf{Top 20}} & \multicolumn{2}{c}{\textbf{Top 100}} \\
                                         & NQ                                  & TQA                                  & NQ                                    & TQA                                      & NQ            & TQA \\
      \midrule
      $l=200$        & 41.1 & 49.7 & 77.9 & 77.9 & 85.4 & 84.0 \\
      $l=500$        & 41.1 & 49.7 & 77.9 & 77.9 & 85.6 & 84.4 \\
      $l=1000$       & 41.1 & 49.7 & 77.9 & 77.9 & \textbf{85.7} & \textbf{84.5} \\
      $l=2000$       & 41.1 & 49.7 & 77.9 & 77.9 & \textbf{85.7} & \textbf{84.5} \\
      \bottomrule
    \end{tabular}
    \caption{Top-1, top-20, and top-100 recall of our model with $l \in \{200, 500, 1000\}$ on test sets.}
   \label{tb:number-of-candidates}
  }
\end{table}

\begin{table}[t]
  \centering
  \small{
    \begin{tabular}{l|ccc}
      \toprule
      \textbf{Configuration}    & \textbf{Top 1} & \textbf{Top 20} & \textbf{Top 100} \\
      \midrule
      Cross entropy loss        & 28.6 & 67.8 & 79.8 \\
      \midrule
      Ranking loss $\alpha=0.0$ & 39.8 & 76.4 & 84.0 \\
      Ranking loss $\alpha=1.0$ & 40.0 & 76.5 & 84.0 \\
      Ranking loss $\alpha=2.0$ & 39.8 & \textbf{76.7} & \textbf{84.1} \\
      Ranking loss $\alpha=4.0$ & \textbf{40.3} & \textbf{76.7} & 84.0 \\
      \bottomrule
    \end{tabular}
    \caption{Top-1, top-20, and top-100 recall of our model with the various settings of the loss function $\mathcal{L}_\text{cand}$ evaluated on the validation set of the NQ dataset.}
   \label{tb:binary-loss}
  }
\end{table}

\section{Effects of Scaling Parameter}
\label{sec:effects-gamma}

To investigate how the scaling parameter, $\gamma$, affects the performance, we test the performance of our model using various $\gamma$ values, where $\gamma \in \{0.025, 0.05, 0.1, 0.2\}$.
The retrieval performance on the validation set of the NQ dataset is shown in Table \ref{tb:experiments-gamma}.
Overall, the scaling parameter has a minor impact on the performance.
We select $\gamma = 0.1$ because of its enhanced performance.

\section{Effects of Number of Candidate Passages}
\label{sec:effects-num-candidates}

We report the performance of our model with the varied number of candidate passages $l$ in Table \ref{tb:number-of-candidates}.
Overall, BPR achieves similar performance in all settings.
Increasing the number of candidate passages slightly improves the top-100 performance until it reaches $l=1000$.

\section{Effects of Loss of Task \#1 with Various Settings}
\label{sec:effects-binary-loss}

We investigate the effects of using various settings of the loss function $\mathcal{L}_\text{cand}$ in Eq.\eqref{eq:ranking-loss}.
Instead of using the ranking loss, we test the performance with the cross-entropy loss, similar to Eq.\eqref{eq:nll-loss}, and $\mathbf{\tilde{h}}_q$ and $\mathbf{\tilde{h}}_p$ are used instead of $\mathbf{e}_q$ and $\mathbf{e}_p$, respectively.
Furthermore, we also test how the parameter $\alpha$ affects the performance.
As shown in Table \ref{tb:binary-loss}, the cross-entropy loss clearly performs worse than the ranking loss.
Furthermore, a change in the parameter $\alpha$ has a minor impact on the performance.
Here, we select the ranking loss with $\alpha=2.0$ because of its enhanced performance on the top-20 and top-100 performance.

\end{document}